\documentclass{article}
\usepackage{ijcai26}

\usepackage{times}
\usepackage{soul}
\usepackage{url}
\usepackage[hidelinks]{hyperref}
\usepackage[utf8]{inputenc}
\usepackage[small]{caption}
\usepackage{graphicx}
\usepackage{xcolor}
\usepackage{amsmath}
\usepackage{amsthm}
\usepackage{booktabs}
\usepackage{algorithm}
\usepackage{algorithmic}
\usepackage[switch]{lineno}
\usepackage{amsfonts}
\usepackage{bbm}
\usepackage{multirow}
\usepackage{makecell}
\usepackage{tabularx}
\usepackage{booktabs}

\title{TS-PEFT: Unveiling Token-Level Redundancy in Parameter-Efficient Fine-Tuning}

\author{
Dabiao Ma\thanks{Equal contribution.}
\and
Ziming Dai\footnotemark[1]\and
Zhimin Xin\footnotemark[1]\And
Shu Wang\and
Jian Yang\and
Haojun Fei
\\
\affiliations
Qfin Holdings, Inc.\\
\emails
\{madabiao-jk,daiziming-jk,xinzhimin-jk,wangshu1-jk,wangye3-jk,zhangchulan-jk\}@qifu.com
}

\begin{document}

\maketitle
\begin{abstract}
    Current Parameter-Efficient Fine-Tuning (PEFT) methods typically operate under an implicit assumption: Once a target module is selected, every token passing through it contributes equally to the downstream task and requires a parameter update.  In this paper, we challenge this convention by revealing a pervasive token-level redundancy in the fine-tuning of large models (LMs). We propose TS-PEFT, a theoretical framework utilizing proximal optimization that acts as a dynamic probe to identify token-level redundancy during the fine-tuning process.  Extensive experiments demonstrate that indiscriminately updating all tokens is not only computationally superfluous but often introduces optimization noise.  Surprisingly, by discarding 30\%-70\% of token updates, TS-PEFT consistently matches or exceeds the performance of dense baselines such as LoRA, DoRA. Our in-depth analysis shows that the learned token-level sparsity is a superior indicator of module importance compared to traditional weight criteria, providing a novel data-driven perspective on the intrinsic adaptation mechanism of LMs.
\end{abstract}

\section{Introduction}

Large models (LMs) are demonstrating astonishing capabilities in Natural Language Processing (NLP)~\cite{yao2024survey,guo2025deepseek} and Computer Vision (CV)~\cite{croitoru2023diffusion,liu2024sora}.  To effectively use these large-scale models in downstream tasks, the dominant paradigm is Parameter-Efficient Fine-Tuning (PEFT)\cite{hu2023llm,hanparameter}.  In these approaches, prompt-based methods such as Prefix-Tuning~\cite{li2021prefix} and Prompt-Tuning~\cite{lester2021power} train additional input tokens to change model behavior and low-rank adaptation (LoRA)~\cite{hu2022lora} and its variants~\cite{zhang2023adaptive,liu2024dora,liu2024dora} change internal representations by injecting a small number of trainable parameters or corridors, i.e. low-rank components, into the base model. These methods effectively decrease trainable parameters while maintaining similar, or comparable, performance to full parameter fine-tuning.

However, despite their distinct architectural designs, these methods share a common and implicit assumption that they all employ dense updates.  That is, once a module like a self-attention layer is used for fine-tuning, the learnable modification is applied indiscriminately to every token position in the input sequence. Formally, for an input sequence $X=\{x_i\}_{i=1}^{T}$ and a frozen weight matrix $W_0$, the hidden state at position $i$ is usually written as:
\begin{equation}
    h_i = W_0 x_i + \operatorname{M}(x_i),
    \label{eq:1}
\end{equation}
where $\operatorname{M}(\cdot)$ is a PEFT module such as LoRA.

\begin{figure*}[t]
    \includegraphics[width=\textwidth]{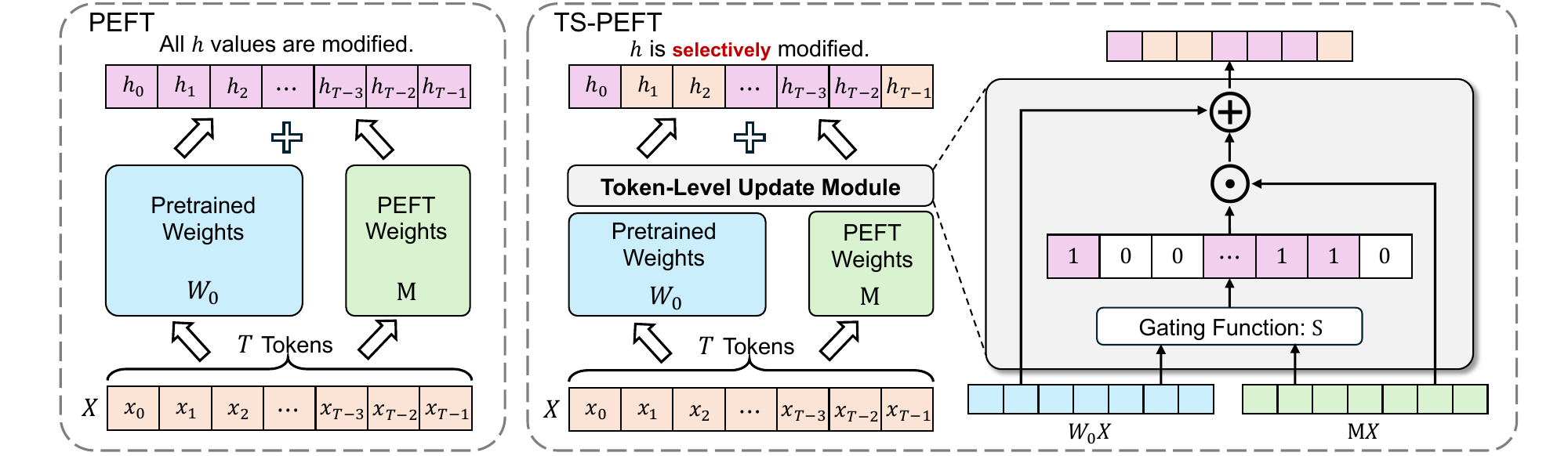}  
    \caption{Comparison between standard PEFT and our TS-PEFT framework.}
    \label{fig:framework}
\end{figure*}

In this paper, we challenge this convention and investigate a fundamental scientific question: \textit{\textbf{Is it necessary to apply fine-tuning updates to all token positions?}}  Given the strong representational power of pretrained base models, we assume that for a specific downstream sample, only a subset of tokens requires task-specific adaptation, while others retain sufficient information from the pretrained weights.  Under this assumption, the standard dense update strategy is suboptimal, primarily because it introduces optimization noise that forces the model to fit gradients on irrelevant tokens, distorting the well-learned features.

Recent works like CFT~\cite{ruan2025enhancing} have demonstrated that masking non-critical tokens during training can improve reasoning. This finding corroborates our intuition that, just as imposing a uniform penalty on all tokens in pretraining may hurt generalization, updating all tokens indiscriminately in PEFT similarly introduces redundancy and noise. To systematically verify this assumption and quantify the potential redundancy, we propose \textbf{T}oken-\textbf{S}elective \textbf{PEFT} (TS-PEFT). Unlike prior approaches, TS-PEFT serves as a learnable probe that dynamically decides whether to update a token or keep its representation frozen. We formulate this selection process as a sparsity-regularized optimization problem, employing proximal optimization and an Adam-style adaptive thresholding mechanism to ensure stable, end-to-end training of the gating decisions.

Our empirical findings unveil a pervasive token-level redundancy. Across many benchmarks,  we observe that nearly 30\%-70\% of token updates are actually redundant. Crucially, we prove that indiscriminately updating these tokens is detrimental. As evidenced by our analysis, forcing updates on these redundant positions leads to a ``performance cliff'', confirming that redundancy in PEFT acts as harmful noise. By filtering out this noise, TS-PEFT consistently matches or exceeds the performance of dense baselines (e.g., LoRA, DoRA) while utilizing significantly fewer active parameters.

Furthermore, our analysis exposes a profound connection between token sparsity and module importance.  We discover that the learned sparsity pattern is not random. The modules that exhibit low sparsity, i.e., requiring frequent updates, are significantly more important to the downstream task than those with high sparsity. This insight allows token-level sparsity to serve as a superior indicator for module selection compared to traditional weight-norm-based metrics. In summary, our contributions are as follows:
\begin{itemize}
    \item \textbf{Unveiling Redundancy:} We identify and quantify that in standard PEFT, nearly 30\%-70\% of token updates are redundant. We empirically demonstrate that indiscriminately updating these tokens is suboptimal and introduces optimization noise.

    \item \textbf{Methodological Framework:} We propose TS-PEFT, a theoretical framework that employs proximal optimization to dynamically learn token-selective strategies in an end-to-end manner.

    \item \textbf{Mechanism Insight:} We reveal that token-level sparsity provides a strong signal for module importance, offering a novel data-driven perspective on the intrinsic adaptation mechanism of LMs.
\end{itemize}

\section{Related Work}

\textbf{Low-Rank Adaptation.} Low-rank adaptation and its derivatives have demonstrated strong potential in reducing the number of trainable parameters while maintaining downstream performance. LoRA~\cite{hu2022lora} introduces two low-rank trainable matrices $A$ and $B$ to approximate weight updates, thereby avoiding direct modification of full weight matrices in large pretrained models. After training, these matrices can be merged back into the base model without incurring additional inference cost. Based on this concept, AdaLoRA~\cite{zhang2023adaptive} adaptively redistributes the effective rank across different layers to better meet task-specific requirements. DoRA~\cite{liu2024dora} refines the decomposition of weight updates to improve the expressiveness and stability of low-rank adaptation, and VeRA~\cite{kopiczkovera} utilizes shared low-rank bases to further reduce the parameter overhead for each task. However, all these methods implicitly assume that once a layer is selected as the PEFT target, every token position passing through that layer should receive an update. That is to say, they perform parameter compression at the module level while still maintaining dense updates along the positional dimension. This may introduce unnecessary redundant modifications and fails to explicitly characterize or exploit potential token-level redundancy.

\noindent \textbf{Sparsity.} The methods of sparsity guidance have been widely used to improve the efficiency and interpretability of large language models (LLMs). By reducing the number of active parameters and concentrating computation on core components, these approaches not only make models more efficient but also offer clearer insights into their decision-making processes. Mixture-of-experts (MoE) models, such as Switch Transformers~\cite{fedus2022switch} and Mixtral 8x7B~\cite{jiang2024mixtral}, employ sparse activation to keep the computational cost roughly constant even with a very large number of parameters. Recently, several works have combined MoE architectures with LoRA-style adapters to further enhance performance~\cite{chen2024llava,liu2024moe,zadouripushing}. Another typical application of sparsity is model pruning, where less influential parameters are removed while preserving important ones to maintain accuracy under lower computational budgets~\cite{ma2023llm,xubesa,xiasheared}. \cite{tan2024sparsity} further introduces sparsity-guided techniques as a tool for interpreting the behavior of LLMs. 

We observe that these methods all utilize redundancy at different levels. MoE-based approaches typically target modular-level sparsity, whereas pruning focuses on parameter-level sparsity. In contrast, our work explores token-level sparsity within the context of PEFT, introducing a gating mechanism to selectively skip updates for tokens requiring no task-specific adaptation. Therefore, our approach is not constrained by these architectures, and theoretically it can be directly embedded as a detachable module into all LoRA variants, including LoRA-MoE, thereby further eliminating redundant computations.

\section{Method}

This section introduces the overall theoretical framework and optimization process of TS-PEFT. We first formalize the LoRA-style PEFT procedure and introduce a token-level control function $\operatorname{S}$ to determine whether a PEFT update should be applied at each position. Then, we formulate the learning of the scalar threshold $\tau$ in $\operatorname{S}$ as an optimization problem with a sparsity regularization term. Next, to address the non-differentiability of $\operatorname{S}$ with respect to $\tau$ and the issue that gradients are nearly zero at most positions, we derive a computable approximate gradient. Finally, we combine this approximate gradient with momentum and adaptive learning rate mechanisms so that it can be seamlessly integrated into the Adam optimizer, and we provide the overall algorithmic description of the TS-PEFT training process.

\subsection{Token-level Selective PEFT Formulation}

As shown in Eq.(\ref{eq:1}), once a layer is selected as a PEFT target layer, all token positions passing through that layer will uniformly receive the increment introduced by $\operatorname{M}(x_i)$. To break this assumption, we introduce a token-level gating function $\operatorname{S}$ on top of Eq.(\ref{eq:1}) to determine whether a PEFT update is actually applied at each position. Specifically, the output $h_i$ at position $i$ is rewritten as:
\begin{equation}
\label{eq:2}
    h_i = 
    \begin{cases}
        W_0x_i & \text{if } \operatorname{S}(W_0x_i, \operatorname{M}(x_i)) = 0, \\
         W_0x_i + \operatorname{M}(x_i) & \text{if } \operatorname{S}(W_0x_i, \operatorname{M}(x_i)) = 1.
    \end{cases}
\end{equation}
Here, $\operatorname{S}(\cdot) \in \{0,1\}$ serves as a gating variable. When $\operatorname{S} = 1$, the update is enabled at that position, while when $\operatorname{S} = 0$, the position retains the base model output. To instantiate Eq.(\ref{eq:2}), we focus on the relative magnitude $r_i$ of the PEFT update with respect to the base model output at each position $i$: $r_i = \|  \operatorname{M}(x_i) \|_2 / \|  W_0x_i \|_2$. This is because the feature norm values at different layers vary significantly. The relative values enable each layer to adaptively adjust according to its own weight change ratio. Intuitively, a larger $r_i$ means that the PEFT update induces a stronger perturbation relative to the base model output at that position and is more likely to be relevant to the task, whereas a small $r_i$ indicates that the modification is minor and can be treated as negligible. Therefore, we introduce a learnable scalar threshold $\tau$ for each target layer and define the control function accordingly:
\begin{equation}
    \operatorname{S}(W_0, \operatorname{M}(x_i)) = 
     \begin{cases}
        0 & \text{if } r_i < \tau,\\
        1 &  \text{if }  r_i \geq \tau.
    \end{cases}
\end{equation}
In other words, if the relative magnitude at a certain position is not lower than $\tau$, we consider the update at that position worth retaining. Otherwise, the update for that position is skipped. This will form a token-level sparse update pattern along the sequence dimension within each selected PEFT module, and the threshold $\tau$ controls the overall sparsity level.

However, this threshold-based step gating introduces two training challenges: First, $\operatorname{S}$ is a non-smooth step function with respect to $\tau$, so standard backpropagation provides almost no useful gradients. Second, we want $\operatorname{S}$ to activate PEFT updates only on a subset of token positions while keeping the remaining ones identical to the base model. This requires explicit control of the overall sparsity level during threshold learning. The next subsections tackle these two issues via a proximal objective and a computable gradient approximation.

\subsection{Proximal Optimization for the Threshold}
Due to the aforementioned training difficulties, directly using first-order optimization methods cannot effectively update $\tau$. Thus, in order to ensure that the threshold can be updated stably during the training process, we formulate its learning process as a proximal optimization problem, enabling it to follow the changes in the original task loss and explicitly favor a sparser update pattern. Specifically, our objective function is defined as follows:
\begin{equation}
    \min \mathcal{L}(\tau) = \ell(\tau) + \lambda\sum_{i} \mathbbm{1}(r_i \geq \tau),
\end{equation}
where $\ell(\tau)$ denotes the loss of the target task, and the second term penalizes the number of excessively activated tokens to encourage a sparser update. $\mathbbm{1}(\cdot)$ is an indicator function, and $\lambda$, as a hyperparameter, is used to balance the sparsity level. 

We employ proximal optimization to derive the update rule for parameter $\tau$ within a stochastic gradient descent (SGD) framework~\cite{parikh2014proximal}. Given the threshold $\tau^k$ at the $k$-th iteration, the update $\tau^{k+1}$ can be written as:
\begin{equation}
\label{eq:5}
\underset{\tau}{\mathrm{argmin}} \left[  \ell(\tau^k) + \frac{\partial \ell}{\partial \tau}(\tau - \tau^k) + \frac{1}{2\alpha_k} \left| \tau - \tau^k \right|^2 + \lambda P(\tau) \right],
\end{equation}
where $\alpha_k$ is the step size. For convenience in the subsequent derivation, we set $\sum_{i} \mathbbm{1}(r_i \geq \tau)$ to be $P(\tau)$. It can be seen that this subproblem consists of three components: a first-order approximation of the loss at $\tau^k$, a proximal term that constrains the magnitude of the update, and a sparsity regularization term that favors fewer activated positions.

To solve Eq.(\ref{eq:5}), we need to compute $\partial \ell / \partial \tau$ and $\partial P / \partial \tau$. Thus, we first expand the output of Eq.(\ref{eq:2}) along the hidden dimension. Let the hidden representation of the $i$-th token at dimension $j \in \{0, \dots, H-1\}$ be denoted as:
\begin{equation}
\label{eq:6}
h_{ij} = \left[ W_0x_i \right]_j + \left[ \operatorname{M}(x_i)\ \right]_j * \mathbbm{1} \left( r_i \geq \tau \right),
\end{equation}
where $H$ is the hidden size, and $[\cdot]_j$ denotes the $j$-th component of a vector. By backpropagation, we finally get:
\begin{align}
\frac{\partial \ell}{\partial \tau} = \sum_{i} \mu_i \frac{\partial }{\partial \tau}\mathbbm{1} \left( r_i \geq \tau \right), \label{eq:7} \\
\frac{\partial P}{\partial \tau} = \sum_{i} \lambda \frac{\partial  }{\partial \tau}\mathbbm{1}\left( r_i \geq \tau \right),
\end{align}
where $\mu_i$ aggregates the overall influence of the PEFT update at position $i$ on the loss:
\begin{equation}
\mu_i = \sum_{j=0}^{H-1} \frac{\partial \ell}{\partial h_{ij}} \left[ M(x_i)\ \right]_j.
\label{eq:9}
\end{equation}

\subsection{Gradient Approximation for the Step Function}
In the previous subsection, we obtain the derivative expressions of the loss function and the sparsity penalty term with respect to the threshold $\tau$. However, all these derivatives depend on the partial derivative of the step gating $\mathbbm{1}(r_i \ge \tau)$ with respect to $\tau$, and its numerical value is almost zero everywhere, making it impossible to directly perform gradient updates. Therefore, to enable stable learning of the threshold through backpropagation, this section constructs a computable gradient approximation based on the perspective of proximal optimization to replace the non-differentiable derivative of the indicator function.

\begin{algorithm}[t]
\caption{Forward and Backward}
\label{alg:fb}
\begin{algorithmic}[1]
    \STATE \textbf{Input:} $\operatorname{M}$, $\alpha_k$, $s$, $\lambda$, $\beta_1$, $\beta_2$, $\tau^k$.
    \STATE \textbf{Output:} $\tau^{k+1}$, updated parameters of $\operatorname{M}$.
    \STATE \textbf{Forward Pass:}
    \STATE \quad Forward as in Eq.(\ref{eq:2}).
    \STATE \textbf{Backward Pass:}
    \STATE \quad \textbf{Step 1:} Perform the usual backward step to update parameters of $\operatorname{M}$.
    \STATE \quad \textbf{Step 2:} 
    \STATE \quad \quad Compute $\mu_i$ as in Eq.(\ref{eq:9}).
    \STATE \quad \quad Compute $g_k$ as in Eq.(\ref{eq:15}).
     \STATE \quad \quad Update $\tau^{k+1}$ as in Eq.(\ref{eq:20}).

\end{algorithmic}
\end{algorithm}

A straightforward approximation, inspired by~\cite{bengio2013estimatingpropagatinggradientsstochastic}, is to treat the gradient of the step function as a constant $-s, s>0$, that is:
\begin{equation}
\frac{\partial }{\partial \tau} \mathbbm{1} \left( r_i \geq \tau \right)\approx -s.
\label{eq:10}
\end{equation}
Substituting this approximation value into Eq.(\ref{eq:7})-(\ref{eq:9}) yields the rough gradients of the loss term and the sparsity term with respect to the threshold:
\begin{align}
\frac{\partial \ell}{\partial \tau} &\approx -s\sum_{i} \mu_i , \label{eq:11} \\
\frac{\partial P}{\partial \tau} &\approx -s\sum_{i} \lambda \label{eq:12}.
\end{align}
Eq.(\ref{eq:10}) is a rather rough approximation, which causes highly oscillatory
updates of $\tau$ and results in poor model performance. We believe this is because Eq.(\ref{eq:10}) should be 0 for certain value domains, but for the update, we set it to $-s$. Therefore, to limit this effect, we add indicator conditions. We modify Eq.(\ref{eq:11}) and Eq.(\ref{eq:12}) as follows:
\begin{align}
    \frac{\partial \ell}{\partial \tau} = -s&\sum_{i} \mathbbm{1}\left[\mathbbm{1} \left(\mu_i \geq 0\right) =  \mathbbm{1} \left(r_i \geq \tau\right)\right]\mu_i , \label{eq:13} \\
    &\frac{\partial P}{\partial \tau} = -s\sum_{i} \mathbbm{1}\left(r_i  \geq \tau\right)\lambda \label{eq:14}.
\end{align}
The constraints can be understood as follows: Whenever $\tau$ is either large enough ($\mu_i{\geq}0$, $r_i{<}\tau$) or small enough ($\mu_i{\leq}0$, $r_i{\geq}\tau$) to render the update of $\tau$ unnecessary, Eq.(\ref{eq:13}) zeros $\partial \ell /\ \partial \tau$. Similarly, when $\tau$ is large enough ($r_i{<}\tau$, $\lambda$ is a positive constant),  Eq.(\ref{eq:14}) nullifies $\partial P /\ \partial \tau$.

Combining both parts, the approximate gradient for updating the threshold can be unified as:
\begin{equation}
    g_k = \sum_{i}  \left\{\mathbbm{1}\left[\mathbbm{1} \left(\mu_i \geq 0\right) =  \mathbbm{1} \left(r_i \geq \tau\right)\right]\mu_i + \mathbbm{1}\left(r_i  \geq \tau\right) \lambda \right\}.\label{eq:15}
\end{equation}
Finally, given the learning rate $\alpha_k$, the threshold is updated as $\tau^{k+1} =\tau^k +  \alpha_ks \cdot g_k$, where $-\alpha_ks$ can be regarded as the learning rate.
\subsection{Adam-style Threshold Update and Overall Training Procedure}

In fact, for the threshold $\tau$, its approximate gradient $g_k$ is highly noisy and varies in scale across training steps. The contributions of tokens change significantly between batches, and the magnitude of $g_k$ can fluctuate during optimization. Directly applying plain gradient descent to $\tau$ often leads to oscillations and can slow down or make the convergence unstable. To mitigate this, we adopt a rule similar to Adam to update, which combines momentum and adaptive learning rate~\cite{kinga2015method}.

At the $k$-th iteration, we use the approximate gradient $g_k$ obtained in the previous subsection to update the first and second moments:
\begin{align}
    m_k &\leftarrow \beta_1m_{k-1} + (1-\beta_1)g_k, \label{eq:16}\\
    v_k &\leftarrow \beta_2v_{k-1} + (1-\beta_2)g_k^2, \\
    \hat{m}_k &\leftarrow m_k / (1 - \beta_1^k), \\ 
    \hat{v}_k &\leftarrow v_k / (1 - \beta_2^k), 
\end{align}
where $\beta_1$ controls the contribution of past gradients and $\beta_2$ governs the decay rate of squared gradients. $\hat m_k$ and $\hat v_k$ denote the bias-corrected first- and second-moment estimates used in the Adam update. Finally, the threshold is updated as:
\begin{equation}
    \tau^{k+1} \leftarrow \tau^k + \alpha_ks\frac{\hat{m}_k}{\sqrt{\hat{v}_k} + \epsilon},\  \text{with } \tau^{0} = 0,
    \label{eq:20}
\end{equation}
where $s$ is the gradient scaling factor, and $\epsilon$ is a minuscule constant to prevent division by zero. We empirically observe that the adaptive moment estimation (Eq.(\ref{eq:16})-Eq.(\ref{eq:20})) is critical for convergence. Ablation studies in Appendix A.4 demonstrate that removing this component leads to severe threshold oscillation and training divergence, resulting in significant performance degradation on sensitive tasks.

\begin{table*}[t]
\centering
\small
\setlength{\tabcolsep}{3pt}
\begin{tabular}{lcccccccccc}
\toprule
  & \textit{Hyperparams} & PIQA & BoolQ & HellaSwag & WinoGrande & SIQA & OBQA & ARC-e & ARC-c & Avg  \\
\midrule
LoRA   &  \multirow{3}{*}{\makecell{$s=4\mathrm{e}{-5}$ \\ $\lambda=4.5\mathrm{e}{-5}$}}
  & 88.5 & 64.2 & 95.3 & \textbf{85.5} & 80.8 & 85.4 & \textbf{90.3} & \textbf{79.9} & 83.7  \\
TS-LoRA   &  & \textbf{88.6} & \textbf{70.1} & \textbf{95.5} & 84.4 & \textbf{82.3} & \textbf{85.8} & 90.1 & 79.4 & \textbf{84.5}  \\
\cmidrule{3-11}
Sparsity(\%) &   & 58.8 & 55.6 & 57.1 & 54.0 & 54.9 & 56.8 & 59.2 & 60.0 & 57.1 \\
\midrule
DoRA   &  \multirow{3}{*}{\makecell{$s=4\mathrm{e}{-5}$\\ $\lambda=1\mathrm{e}{-5}$}}   & 87.5 & 75.0 & 95.5 & 85.5 & 80.1 & 85.8 & 90.8 & 79.5 & 85.0 \\
TS-DoRA  &   & \textbf{88.8} & \textbf{75.2} & 95.5 & \textbf{87.1} & \textbf{80.2} & \textbf{86.6} & \textbf{91.2} & \textbf{80.1} & \textbf{85.6} \\
\cmidrule{3-11}
Sparsity(\%) &   & 50.0 & 47.9 & 50.2 & 47.1 & 47.6 & 48.3 & 49.6 & 50.0 & 48.8 \\
\midrule
AdaLoRA   &  \multirow{3}{*}{\makecell{$s=4\mathrm{e}{-5}$\\  $\lambda=1\mathrm{e}{-4}$}}  & \textbf{88.8} & 74.2 & 95.5 & 85.2 & 79.6 & 85.4 & \textbf{91.2} & 79.5 & 84.9\\
TS-AdaLoRA   &   & 88.4 & \textbf{75.5} & \textbf{95.8} & \textbf{85.6} & \textbf{80.9} & \textbf{86.2} & 90.1 & \textbf{79.9} & \textbf{85.3} \\
\cmidrule{3-11}
Sparsity(\%) &   & 74.3 & 76.0 & 65.0 & 70.8 & 68.9 & 72.8 & 75.5 & 75.9 & 67.7  \\
\bottomrule
\end{tabular}
\caption{Evaluation scores of LLaMA3.1-8B on CSR benchmarks; the last row in each block shows the sparsity of TS-PEFT.}
\label{tab:common}
\end{table*}

Combining the above components, the training process of TS-PEFT can be summarized as follows: In each training step, during the forward propagation phase, we perform gating based on each token’s relative update magnitude $r_i$ and the threshold $\tau$ of the corresponding layer, applying updates only at positions that satisfy the condition while keeping the base model output unchanged at all other positions. Then, standard backpropagation is used to update all PEFT parameters, while the base model parameters remain frozen throughout. On this basis, we compute $\mu_i$ to construct the approximate gradient $g_k$ for the threshold, and independently update the threshold $\tau$ of each layer using Adam-style first- and second-moment estimates. The full training procedure is presented in Algorithm \ref{alg:fb}.

\section{Experiments}

\subsection{Experimental Setup}

We evaluate TS-PEFT in multiple representative scenarios: Natural Language Understanding (NLU), Commonsense Reasoning (CSR), Visual Instruction Tuning (VIT), and Natural Language Generation (NLG). Due to space limitations, we present the detailed results and analysis for NLG in Appendix A.3. The experimental setup covers the datasets, baseline models, and hyperparameter configurations to ensure reproducibility and comparability of the results.

\paragraph{Datasets and Models.} We select several representative benchmarks in the current PEFT and small-model fine-tuning literature. For CSR, we evaluate on LLaMA-3.1-8B~\cite{dubey2024llama} using PIQA, BoolQ, HellaSwag, WinoGrande, SIQA, OBQA, and ARC-e/ARC-c, covering typical language reasoning scenarios such as physical commonsense, boolean question answering, narrative completion, and scientific QA~\cite{liu2024dora}. For VIT, we evaluate LLaVA-1.5-7B~\cite{liu2023visual,liu2024improved} on ScienceQA, POPE, MMBench, GQA, VisWiz, VQAv2, and VQAT, which are generative language tasks with open-ended answers conditioned on images and textual instructions. For NLU, we build on DeBERTaV3-base~\cite{hedebertav3} and use the GLUE benchmark datasets~\cite{wang2018glue}. The GLUE tasks follow the official evaluation protocol: CoLA uses the Matthews correlation coefficient, STS-B uses correlation coefficients, and all other tasks use accuracy. Additionally, for the NLG tasks (detailed in Appendix), we evaluate Mistral-7B-v0.1~\cite{jiang2023mistral7b} on the Vicuna~\cite{vicuna2023}, LIMA~\cite{zhou2023lima} and Dolly~\cite{DatabricksBlog2023DollyV2} datasets.

\paragraph{Baselines.} In our experiments, we select several mainstream PEFT methods as baselines, covering two major categories: low-rank update methods and adapter-based methods. Specifically, LoRA, DoRA, and AdaLoRA are used to examine whether TS-PEFT can consistently provide additional gains under different low-rank update strategies, while Parallel Adapter~\cite{hu2023llm} is included to evaluate the method’s scalability on adapter-style architectures. For each baseline, we keep its original implementation, hyperparameters, and rank configuration unchanged, and apply TS-PEFT on top of it. This ensures that the comparison is not affected by factors unrelated to the baseline itself and that the results accurately reflect the impact of token-level selective updates.
\paragraph{Training details.}
In all experiments, we freeze the base model and update only the PEFT parameters and the threshold $\tau$. We use the AdamW optimizer, where PEFT is the default learning rate and the effective learning rate of $\tau$ is further modulated by the scaling factor $s$. For text tasks, we follow standard instruction-tuning configurations. For vision–language tasks, we adopt the original training pipeline of the corresponding models. Unless otherwise specified, we set $\alpha_k = 1$, $\beta_{1} = 0.9$, and $\beta_{2} = 0.98$. Each main experiment is repeated at least three times with different random seeds, and we report the mean scores across runs. Additional hyperparameter settings are summarized in Appendix A.6.

\subsection{Experimental Results}

\begin{table*}[t]
\centering
\small
\setlength{\tabcolsep}{3pt}
\begin{tabular}{lccccccccc}
\toprule
  & \textit{Hyperparams} & SQA & POPE & MMBench & GQA & VisWiz & VQAv2 & VQAT & Avg \\
\midrule
LoRA   & \multirow{3}{*}{\makecell{$s=4\mathrm{e}{-5}$\\ $\lambda=2\mathrm{e}{-6}$}}
       & 68.1 & 87.3 & \textbf{64.8} & 63.1 & 46.8 & \textbf{79.1} & 57.3 & 66.6 \\
TS-LoRA &  & \textbf{68.3} & \textbf{87.5} & 63.6 & \textbf{63.3} & \textbf{50.4} & 78.3 & \textbf{57.5} & \textbf{67.0} \\
\cmidrule{3-10}
Sparsity(\%) &   & 67.6 & 59.3 & 62.0 & 59.3 & 59.5 & 59.2 & 62.2 & 61.3 \\
\midrule
DoRA   & \multirow{3}{*}{\makecell{$s=4\mathrm{e}{-5}$\\ $\lambda=1\mathrm{e}{-6}$}}
       & 67.5 & 87.4 & \textbf{65.5} & \textbf{63.0} & 50.5 & \textbf{78.6} & \textbf{57.0} & 67.1 \\
TS-DoRA &  & \textbf{69.4} & \textbf{88.0} & 64.7 & 62.4 & \textbf{54.2} & 77.8 & 55.9 & \textbf{67.4} \\
\cmidrule{3-10}
Sparsity(\%) &   & 70.6 & 62.6 & 65.2 & 62.4 & 62.9 & 63.8 & 65.2 & 64.7 \\
\midrule
AdaLoRA   & \multirow{3}{*}{\makecell{$s=4\mathrm{e}{-5}$\\ $\lambda=5\mathrm{e}{-7}$}}
          & \textbf{68.0} & 87.3 & \textbf{64.3} & 61.7 & 48.9 & 78.3 & \textbf{56.9} & 66.5 \\
TS-AdaLoRA &  & 67.8 & \textbf{87.5} & 63.4 & \textbf{62.3} & \textbf{52.6} & \textbf{78.5} & 55.2 & \textbf{66.8} \\
\cmidrule{3-10}
Sparsity(\%) &   & 60.0 & 50.9 & 54.1 & 50.9 & 51.4 & 50.9 & 53.9 & 53.2 \\
\bottomrule
\end{tabular}
\caption{Evaluation scores of LLaVA-1.5-7B on VIT benchmarks; the last row in each block shows the sparsity of TS-PEFT.}
\label{tab:llava}
\end{table*}

\begin{table*}[t]
\centering
\small
\setlength{\tabcolsep}{3pt}

\begin{tabular}{lcccccccccc}
\toprule
Method      & MNLI & CoLA & QNLI & RTE & MRPC & QQP & SST-2 & STS-B & Avg \\
\midrule
LoRA        & 89.9 & 69.2 & 94.1 & 87.1 & 90.4 & \textbf{92.2} & 95.3 & \textbf{91.8} & 88.8 \\
TS-LoRA     & \textbf{89.9} & \textbf{70.3} & \textbf{94.2} & \textbf{88.1} & \textbf{91.2} & 92.0 & \textbf{95.8} & 91.5 & \textbf{89.1} \\
\cmidrule{2-10}
Sparsity(\%)& 71.6 & 64.1 & 70.6 & 42.8 & 55.4 & 57.1 & 67.6 & 33.9 & 57.9 \\
$\lambda$   & 1e-5 & 6e-6 & 2e-5 & 1e-5 & 1e-5 & 2e-6 & 2e-5 & 8e-5 & -\\
\midrule
AdaLoRA     & 90.4 & 71.1 & \textbf{94.6} & 88.6 & 91.3 & 91.9 & 95.9 & \textbf{91.9} & 89.5 \\
TS-AdaLoRA  & \textbf{90.5} & \textbf{71.3} & 94.5 & \textbf{89.0} & \textbf{91.6} & 91.9 & \textbf{96.1} & 91.6 & \textbf{89.6} \\
\cmidrule{2-10}
Sparsity(\%)& 55.2 & 53.1 & 68.1 & 62.7 & 71.7 & 54.7 & 56.9 & 68.3 & 61.3 \\
$\lambda$   & 9e-6 & 5e-6 & 4.5e-7 & 2e-6 & 2e-8 & 5e-7 & 2e-6 & 5e-6 & -\\
\midrule
Adapter     & \textbf{90.4} & \textbf{72.1} & \textbf{94.2} & 86.6 & 90.0 & \textbf{92.1} & 95.6 & \textbf{91.4} & 89.1 \\
TS-Adapter  & 90.1 & 72.0 & 94.1 & \textbf{86.8} & \textbf{91.4} & 91.8 & \textbf{96.6} & 91.3 & \textbf{89.3} \\
\cmidrule{2-10}
Sparsity(\%)& 56.4 & 59.4 & 65.5 & 62.7 & 55.1 & 58.0 & 56.2 & 67.7 & 60.1 \\
$\lambda$   & 5e-7 & 4.5e-7 & 4.5e-7 & 2e-6 & 2e-8 & 5e-7 & 2e-6 & 5e-6 & -\\
\bottomrule
\end{tabular}
\caption{Evaluation scores of DeBERTaV3-base on GLUE benchmarks; the ``Sparsity(\%)" rows show the sparsity of TS-PEFT, and the ``$\lambda$" rows list task-wise thresholds ($s$ = 1e-4).}
\label{tab:glue}
\end{table*}

\noindent \textbf{Commonsense Reasoning.} Table \ref{tab:common} reports the results of various PEFT methods and their TS-PEFT variants on the CSR tasks. For a fair comparison, we set the rank of both LoRA and DoRA to 32, and configure AdaLoRA with an initial rank of 64 and a target rank of 32. Under the same rank settings, we observe that TS-LoRA, TS-DoRA, and TS-AdaLoRA achieve performance comparable to or better than their respective baselines on most datasets. This indicates that selectively skipping updates on a subset of tokens does not weaken downstream performance. Instead, it can help alleviate the noise introduced by redundant updates.

The table also reports the sparsity rate of TS-PEFT, which is the proportion of positions in the input sequence that are \textbf{not} modified by the PEFT output, as defined in Eq.(\ref{eq:21}). Across different baselines and tasks, this proportion remains roughly within the 50\%–70\% range, meaning that only 30\% to 50\% of the tokens actually receive PEFT updates. Even under this degree of sparsity, TS-PEFT is able to match or outperform the original baselines in commonsense reasoning, suggesting that standard PEFT exhibits substantial redundancy along the token dimension, and that threshold-based token-level selective updating provides a more efficient use of PEFT capacity.
\begin{equation}
\label{eq:21}
     \text{Sparsity}(W_0, \operatorname{M}, X) = 1 - \frac{1}{T}\sum_{i}{\operatorname{S}(W_0x_i, \operatorname{M}(x_i))} .
\end{equation}

\noindent \textbf{Visual Instruction Tuning.} For visual instruction tuning, we follow the training setup of \cite{liu2024dora}, configuring LoRA and DoRA with rank 128 and setting the initial and target ranks of AdaLoRA to 256 and 128. Table~\ref{tab:llava} shows the results. In all benchmark tests, TS-LoRA, TS-DoRA, and TS-AdaLoRA achieve consistent improvements of about 0.3–0.4 points, while operating under 50\%–70\% token sparsity. This demonstrates that token-level redundancy is also present in multimodal generative tasks, and selective token updates remain effective, consistent with our observations in CSR.

\noindent \textbf{Natural Language Understanding.} The results are shown in Table~\ref{tab:glue}, where the target ranks of LoRA and AdaLoRA are both set to 4. Overall, TS-LoRA and TS-AdaLoRA achieve average performance that is comparable to or slightly better than their respective baselines, consistent with the trends observed in the previous two task categories. To test whether our approach is tied to a specific low-rank parameterization, we further include Parallel Adapter as a non–LoRA-style PEFT baseline and apply TS-PEFT on top of it to obtain TS-Adapter, using a hidden size of 32. The results show that TS-Adapter performs no worse than the original Adapter, suggesting that TS-PEFT is not bound to LoRA-style weight structures, but can function as a general token-level gating layer that can be stacked on diverse PEFT modules.

TS-PEFT has achieved PEFT updates for only 30\% - 70\% of the tokens in multiple tasks, yet still maintain or even surpass the baseline performance. This clearly reveals the widespread token-level redundant updates in the current PEFT mode, and these updates also introduce noise that affects model performance. By quantifying this intrinsic redundancy, TS-PEFT establishes a clear blueprint for translating theoretical sparsity into substantial physical efficiency gains in the future.

\subsection{Analysis: Token Sparsity Reveals Model Adaptation Mechanism} 
\label{sec:Indepth}

\begin{table*}[t]
\centering
\small
\setlength{\tabcolsep}{3.6pt}      
\begin{tabular}{lcccccccccc}
\toprule
 & PIQA & BoolQ & HellaSwag & WinoGrande & SIQA & OBQA & ARC-e & ARC-c & Avg \\
\midrule
S\_low\_50\%        & \textbf{88.6} & \textbf{75.7} & 95.4 & \textbf{86.9} & 80.6 & \textbf{87.6} & \textbf{90.8} & 79.7 & \textbf{85.7} \\
S\_high\_50\%       & 88.2 & 62.2 & 92.2 & 85.2 & 80.2 & 86.2 & 90.4 & 79.9 & 83.0 \\
norm\_relative\_50\% & 88.4 & 62.2 & 95.4 & 85.0 & 80.5 & 83.6 & 90.1 & \textbf{80.2} & 83.2 \\
norm\_abs\_50\%      & 88.4 & 52.7 & 85.8 & 83.9 & 78.7 & 86.2 & 89.6 & 79.0 & 80.5 \\
random\_50\%        & 88.3 & 73.4 & 91.2 & 85.6 & 80.6 & 85.7 & 89.6 & 78.9 & 84.2 \\
half\_rank\_50\%    & 88.1 & 75.4 & 95.2 & 86.3 & \textbf{81.2} & 85.2 & 89.6 & 78.7 & 84.9 \\
\bottomrule
\end{tabular}
\caption{Performance of LLaMA3.1-8B on the commonsense reasoning task using various selection methods.}
\label{tab:importance_1}
\end{table*}
\begin{figure*}[t]

    \centering
    \includegraphics[width=\textwidth]{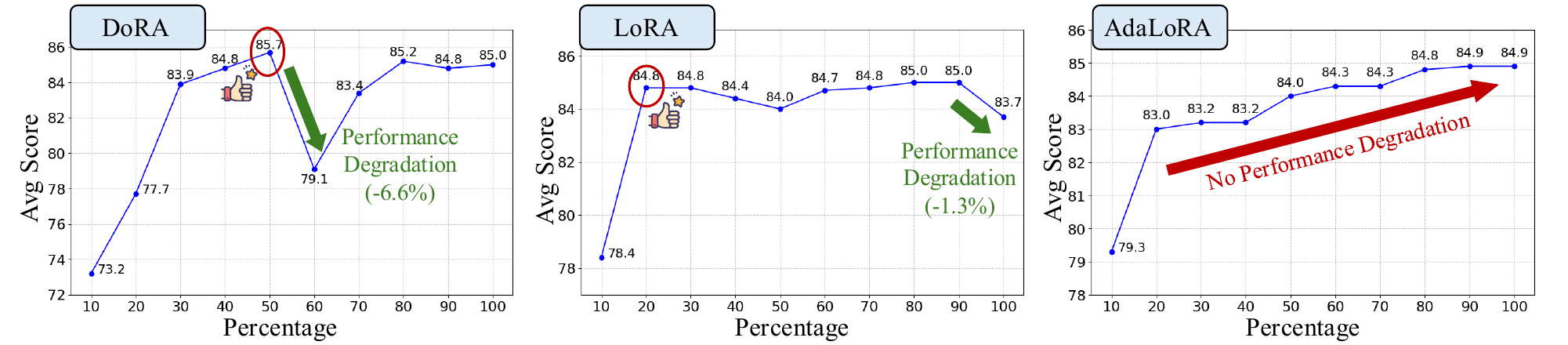}  
    \caption{Performance of LLaMA3.1-8B on CSR benchmarks as the percentage of selected modules varies for each PEFT method. Note the performance drops (e.g., DoRA at 60\%) when redundant, high-sparsity modules are forced to update, indicating noise injection.}
    \label{fig:three_images}
   
\end{figure*}

\noindent \textbf{Sparsity Patterns Reflect Architecture Bias.} There are 32 layers in LLaMA3.1-8B. We compute the sparsity of q\_proj, k\_proj, and v\_proj in each layer and take their average to obtain the sparsity distribution of TS-PEFT across modules (detailed results are provided in Appendix A.5). Overall, the sparsity of TS-LoRA and TS-AdaLoRA is relatively evenly distributed across layers, with standard deviations of 9.8 and 11.4, respectively. In contrast, TS-DoRA exhibits a much higher standard deviation of 41.3, which shows a more polarized pattern. It should be noted that LoRA and AdaLoRA share the same weight structure despite having different rank allocation strategies, whereas DoRA adopts a distinct parameterization by decomposing weights into magnitude and direction. We hypothesize that this structural difference leads to the stronger selectivity observed in TS-DoRA at the module level, naturally raising the question: \textit{\textbf{Are the modules with lower sparsity more critical for downstream tasks?}}

\noindent \textbf{Token-level Sparsity as a Module Importance Indicator.} To examine whether token-level sparsity can be used to identify important modules, we design a set of selection experiments based on the sparsity patterns learned by TS-DoRA. After completing TS-DoRA fine-tuning, we rank all modules by their average token sparsity and construct two main settings under the same parameter budget: (1) S\_low\_50\%, which fine-tunes only the 50\% of modules with the lowest sparsity, and (2) S\_high\_50\%, which fine-tunes only the 50\% of modules with the highest sparsity. 

We compare these sparsity-based selections against several alternatives: norm\_relative\_50\% (selecting the 50\% of modules with the largest average $\lVert \operatorname{M}(x_i)\rVert_2 / \lVert W_0 x_i\rVert_2$), norm\_abs\_50\% (the 50\% with the largest average $\lVert \operatorname{M}(x_i)\rVert_2$), random\_50\% (randomly selecting 50\% of modules, averaged over three runs), and half\_rank\_50\% (using all modules but halving the rank so that the total number of trainable parameters matches the 50\% settings). The results in Table~\ref{tab:importance_1} show that S\_low\_50\% not only significantly outperforms S\_high\_50\% under the same parameter budget, but even surpasses fine-tuning all DoRA modules (whose mean score is 85.0). In contrast, norm-based selections, random choices, and half-rank full-module tuning all yield clearly inferior performance. This indicates that, in our setting, the token-level sparsity patterns learned by TS-DoRA provide a more reliable signal of module importance than simple norm statistics or random selection.

\noindent \textbf{Redundant Updates Introduce Optimization Noise.} To further validate that the identified high-sparsity modules are indeed redundant and potentially harmful, we conduct a cumulative selection analysis. We expand the set of target modules in ascending order of sparsity. Specifically, we add the modules from the key ones to the redundant ones sequentially. As illustrated in Figure \ref{fig:three_images}, the conventional wisdom that ``more trainable parameters imply better performance'' does not hold. For DoRA, performance peaks when only 50\% of the modules are fine-tuned, while LoRA reaches its optimum at 80\%. Remarkably, even utilizing just 20\% of the modules yields results comparable to the optimal settings.

Crucially, we observe distinct performance cliffs when redundant modules are forced to update. For instance, DoRA suffers a sharp performance drop of 6.6\% when the target proportion increases from 50\% to 60\%. Similarly, LoRA experiences a 1.3\% degradation when expanding from 90\% to 100\%. This phenomenon provides compelling evidence that updating modules characterized by high sparsity injects negative interference rather than meaningful adaptation. In contrast, AdaLoRA exhibits greater stability against these low-importance modules, likely due to its dynamic rank allocation mechanism.

\section{Conclusion}
This paper revisits PEFT from a token-level perspective and shows that standard PEFT methods exhibit substantial redundancy by uniformly updating all token positions within selected layers. We propose TS-PEFT, which introduces a learnable threshold gate to selectively apply PEFT updates to only about 30\%–70\% of token positions, yet consistently matches or surpasses strong baselines. Our analyses further show that the resulting token-level sparsity patterns provide a meaningful signal of module importance, enabling more effective module selection under the same parameter budget and offering a theoretical basis for future hardware-software co-design targeting efficient sparse fine-tuning and inference.

\bibliographystyle{named}
\bibliography{ijcai26}

\appendix
\section{Appendix}

\subsection{LLM Usage Statement}
During the preparation of this manuscript, the authors utilized a large language model (LLM) exclusively for grammar and language polishing. All technical content, scientific claims, and experimental results were conceived, derived, and verified solely by the authors.

\subsection{Case Study}

\begin{figure}[htbp]
    \centering
    \includegraphics[width=\linewidth]{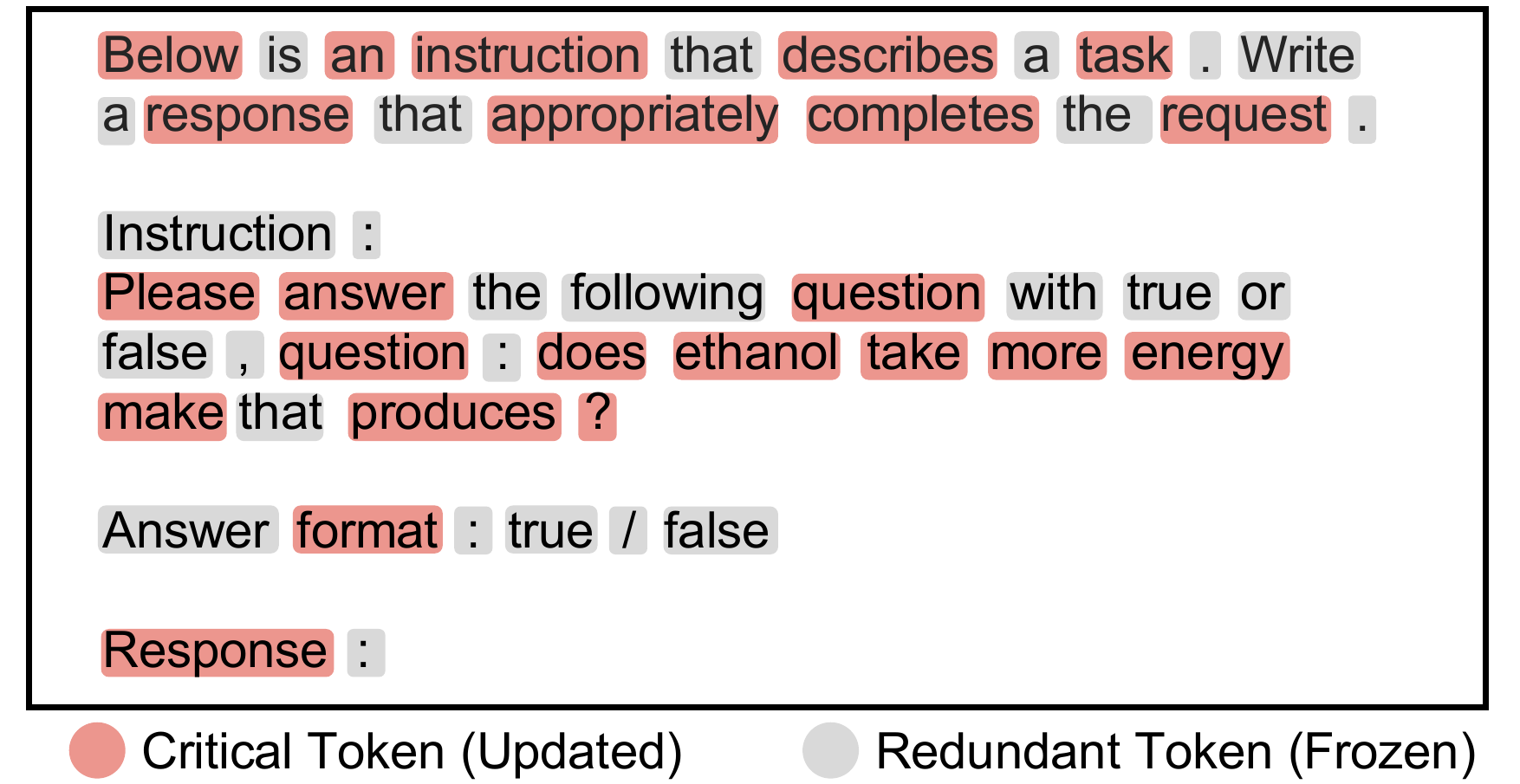}  
    \caption{Visualization of token-level sparsity learned by TS-LoRA on a BoolQ sample.}
    \label{fig:g2}
\end{figure}

To explicitly demonstrate the token selection of TS-PEFT, we visualize the learned token-level sparsity on a representative validation sample from the BoolQ dataset, as shown in Figure \ref{fig:g2}.

By observing the distribution of the sparsity patterns, we find that TS-PEFT exhibits a clear semantic filtering mechanism when processing BoolQ tasks, which is manifested in three aspects:
\begin{itemize}
    \item \textbf{Suppression of High-Frequency Function Words and Syntactic Noise:} The model automatically freezes high-frequency function words and stop words such as ``is'', ``that'', ``the'', and ``:''. These words already possess well-established syntactic functions from the pre-training phase, and forcibly fine-tuning them often introduces unnecessary optimization noise. TS-PEFT effectively preserves the linguistic knowledge of the pre-trained model by skipping these positions.
    \item \textbf{Focus on Critical Domain Entities:} For entity words closely related to the core logic of the question, such as ``ethanol'', ``energy'', and ``produces'', the model chooses to update their parameters. This indicates that TS-PEFT can accurately isolate information carriers and concentrate the optimization budget thereon.
    \item \textbf{Preservation of Generic Pre-trained Concepts:} Interestingly, the model also tends to keep the core label words in the output format, such as ``true'', ``false'', frozen. This implies that the model can reuse the truth-value concepts established during pre-training without re-adapting them during fine-tuning.
\end{itemize}

In conclusion, these qualitative results strongly support the hypotheses presented in the main text. We show that TS-PEFT is capable of acting as a dynamic probe that can autonomously identify and filter out redundant parts in the token level, thereby reducing the noise introduced by fine-tuning in inference tasks.

\subsection{Additional Results}

To further evaluate the generalization capability of TS-PEFT in open-ended text generation scenarios, we extend our experiments to NLG tasks. Specifically, we perform fine-tuning on the Alpaca dataset using the Mistral-7B-v0.1 model and subsequently evaluate performance on three representative benchmarks: Vicuna, LIMA, and Dolly. To measure generation quality comprehensively and objectively, we employed DeepSeek-R1-0325 as an automated evaluator, scoring model outputs on a 10-point scale across four dimensions: accuracy, relevance, fluency, and completeness.

As shown in Table~\ref{tab:nlg}, variants of TS-PEFT consistently outperform their corresponding dense update baselines across all test sets. Specifically, TS-LoRA, TS-DoRA, and TS-AdaLoRA surpass their baselines by average scores of 0.11, 0.10, and 0.12, respectively, while maintaining a token-level sparsity of approximately 75\%, that is, updating only about one-quarter of the tokens. This result not only demonstrates the robustness of TS-PEFT in generation tasks but also provides strong empirical support for the core hypothesis of this paper. That is, in the process of fine-tuning, updating a vast number of non-critical tokens is not only computationally redundant but may even introduce noise, whereas dynamically filtering and skipping these updates via TS-PEFT facilitates the generation of higher-quality and more logically coherent responses.

\begin{table}[t]
\centering
\small
\setlength{\tabcolsep}{3pt}
\begin{tabular}{lccccc}
\toprule
  & \textit{Hyperparams} & Vicuna & LIMA & Dolly & Avg \\
\midrule
LoRA   & \multirow{3}{*}{\makecell{$s=4\mathrm{e}{-5}$\\ $\lambda=1\mathrm{e}{-5}$}}
       & 6.78 & 7.13 & 8.00 & 7.30  \\
TS-LoRA &  & \textbf{6.86} & \textbf{7.20} & \textbf{8.18} & \textbf{7.41}  \\
\cmidrule{3-6}
Sparsity(\%) &   & 76.8 & 78.8 & 76.2 & 77.3  \\
\midrule
DoRA   & \multirow{3}{*}{\makecell{$s=4\mathrm{e}{-5}$\\ $\lambda=6\mathrm{e}{-5}$}}
       & 7.02 & \textbf{6.95} & 7.82 & 7.26  \\
TS-DoRA &  & \textbf{7.09} & 6.83 & \textbf{8.15} & \textbf{7.36}  \\
\cmidrule{3-6}
Sparsity(\%) &   & 75.8 & 76.4 & 73.2 & 75.1  \\
\midrule
AdaLoRA   & \multirow{3}{*}{\makecell{$s=4\mathrm{e}{-5}$\\ $\lambda=6\mathrm{e}{-5}$}}
       & 6.94 & 6.74 & 7.59 & 7.09  \\
TS-AdaLoRA &  & \textbf{6.98} & \textbf{6.76} & \textbf{7.89} & \textbf{7.21}  \\
\cmidrule{3-6}
Sparsity(\%) &   & 72.2 & 77.3 & 75.3 & 74.9  \\
\bottomrule
\end{tabular}
\caption{Evaluation scores of Mistral-7B-v0.1 on NLG benchmarks; the last row in each block shows the sparsity of TS-PEFT.}
\label{tab:nlg}
\vspace{-1mm}
\end{table}

\subsection{The training loss progression with and without the Adaptive Moment Estimation }
\label{apx:a2}

\begin{figure}[t]
    \centering
    \includegraphics[width=\linewidth]{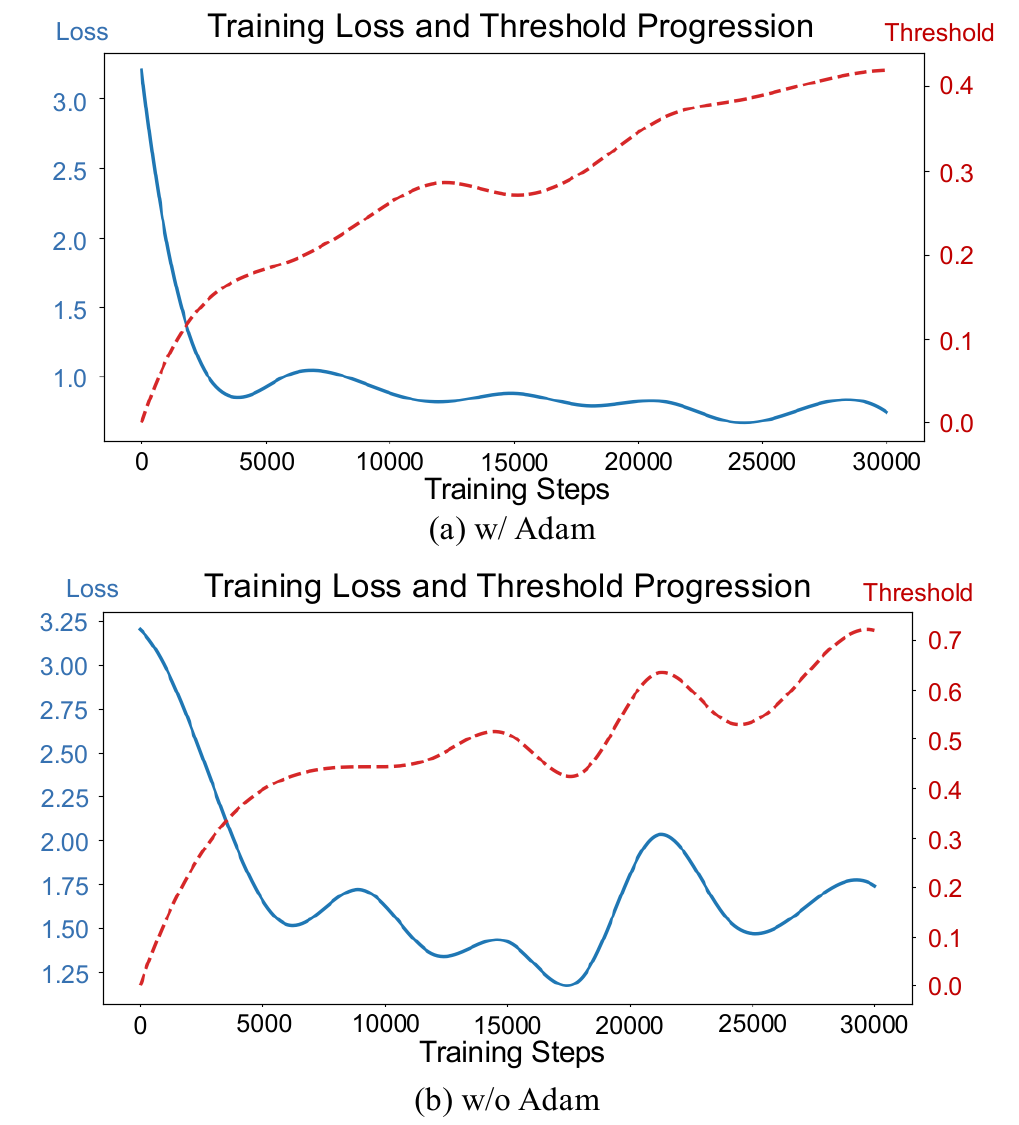}  
    \caption{Training loss progression for layer base\_model.layers.20.self\_attn.v\_proj of LLaMA3.1-8B.}
    \label{fig:adam}
\end{figure}

Figure \ref{fig:adam} illustrates the training dynamics of a single TS-PEFT layer (the v\_proj in layers.20.self\_attn of LLaMA3.1-8B) with and without Adam-style momentum on the threshold $\tau$. In both subfigures, the blue curve shows the training loss, while the red dashed curve tracks the corresponding threshold value over training steps. As shown in Figure \ref{fig:adam} (a), when momentum is enabled, the threshold increases smoothly and gradually saturates around a stable value, while the loss decreases rapidly at the beginning and then oscillates mildly around a low level. This indicates that $\tau$ can adapt to the task in a stable manner without causing large fluctuations in the optimization trajectory.

In contrast, when we remove the adaptive moment estimation and update $\tau$ using plain gradient descent, as shown in Figure \ref{fig:adam} (b), the threshold trajectory becomes highly oscillatory and eventually drifts to a much larger range (above 0.6). The loss curve also exhibits large, persistent oscillations and fails to converge to a similarly low level as in Figure~\ref{fig:adam}(a). This unstable behavior at the layer level is consistent with the degradation observed on downstream CSR benchmarks such as BoolQ. That is, the accuracy drops noticeably compared with the TS-PEFT that uses Adam-style updates for $\tau$. These results provide direct empirical evidence that adaptive momentum estimation is crucial for stabilizing the learning of token-level thresholds in practice.

\subsection{Module-wise average sparsity of LLaMA3.1-8B on the commonsense reasoning task}
\label{apx:a3}
Table \ref{tab:layerwise} reports the detailed module-wise sparsity of TS-LoRA, TS-DoRA, and TS-AdaLoRA on LLaMA3.1-8B for CSR benchmarks. For each transformer layer, that is, indexed from 0 to 31, we measure the token-level sparsity of the PEFT updates in the query, key, and value projections (q\_proj, k\_proj, and v\_proj), and then compute their average, the ``avg'' column in the figure. The last row summarizes the mean sparsity over all layers for each method.

From the table, we observe that TS-LoRA and TS-AdaLoRA exhibit relatively smooth sparsity profiles across layers. Their per-layer averages mostly stay within the 50\%–70\% range, and no single layer dominates the overall sparsity pattern. This matches the moderate standard deviations reported in the main text (9.8 for TS-LoRA and 11.4 for TS-AdaLoRA), suggesting that these two methods tend to distribute token-level updates more evenly over depth. In contrast, TS-DoRA shows a much more polarized pattern: several layers (e.g., layers 0–2, 5–7, 11, 17, 24, 29) have near-100\% sparsity in one or more projections, while a few layers (such as 8, 9, 14, 22) maintain comparatively low sparsity. This resulted in a relatively large standard deviation of 41.3, indicating that TS-DoRA concentrates effective updates into a smaller subset of layers. These fine-grained statistics support our observation in the main paper that the token-level sparsity learned by TS-DoRA can serve as a strong indicator of module importance and motivates the subsequent experiments that select only the low-sparsity (high-importance) modules for fine-tuning.
\begin{table*}[t]
\centering
\small          
\setlength{\tabcolsep}{2.8pt}   

\begin{tabular}{l|cccc|cccc|cccc}
\toprule
 & \multicolumn{4}{c|}{TS-LoRA} & \multicolumn{4}{c|}{TS-DoRA} & \multicolumn{4}{c}{TS-AdaLoRA} \\
\cmidrule(lr){2-5}\cmidrule(lr){6-9}\cmidrule(lr){10-13}
Layer & q\_proj & k\_proj & v\_proj & avg & q\_proj & k\_proj & v\_proj & avg & q\_proj & k\_proj & v\_proj & avg \\
\midrule
0 & 65.6 & 77.6 & 0.0 & 47.7 & 100.0 & 0.0 & 0.0 & 33.3 & - & - & 76.9 & 76.9 \\
1 & 59.7 & 65.5 & 54.8 & 60.0 & 100.0 & 0.0 & 65.0 & 55.0 & 62.3 & 84.5 & 68.4 & 71.8 \\
2 & 62.8 & 64.9 & 51.5 & 59.7 & 100.0 & 100.0 & 57.2 & 85.7 & 57.3 & 75.0 & 81.7 & 71.3 \\
3 & 60.1 & 64.4 & 56.3 & 60.2 & 99.8 & 100.0 & 37.1 & 79.0 & 67.6 & 68.7 & 81.9 & 72.7 \\
4 & 63.7 & 67.6 & 56.6 & 62.6 & 24.6 & 0.0 & 83.0 & 35.9 & 48.2 & 70.2 & 75.8 & 64.7 \\
5 & 57.9 & 62.6 & 54.8 & 58.4 & 0.0 & 0.0 & 30.9 & 10.3 & 49.6 & 86.4 & 80.9 & 72.3 \\
6 & 62.4 & 60.6 & 48.4 & 57.1 & 100.0 & 0.0 & 62.9 & 54.3 & 70.7 & 84.8 & 66.1 & 73.9 \\
7 & 61.3 & 65.9 & 51.2 & 59.5 & 0.0 & 0.0 & 82.8 & 27.6 & 59.8 & 86.6 & 77.2 & 74.5 \\
8 & 60.8 & 65.8 & 50.4 & 59.0 & 70.5 & 57.1 & 60.2 & 62.6 & 54.0 & 93.6 & 60.7 & 69.4 \\
9 & 65.9 & 51.1 & 53.1 & 56.7 & 0.0 & 11.6 & 47.4 & 19.7 & 52.0 & 72.4 & 61.2 & 61.9 \\
10 & 57.7 & 58.7 & 52.8 & 56.4 & 0.0 & 0.0 & 41.3 & 13.8 & 51.3 & 82.5 & 64.2 & 66.0 \\
11 & 58.5 & 57.5 & 42.3 & 52.8 & 91.7 & 77.2 & 76.1 & 81.7 & 54.2 & 86.6 & 65.0 & 68.6 \\
12 & 51.9 & 68.2 & 49.9 & 56.7 & 0.0 & 0.0 & 2.1 & 0.7 & 50.7 & 87.9 & 56.0 & 64.9 \\
13 & 58.9 & 68.3 & 51.2 & 59.5 & 90.2 & 90.0 & 35.1 & 71.7 & 61.3 & 73.4 & 56.1 & 63.6 \\
14 & 57.1 & 60.6 & 48.9 & 55.6 & 0.0 & 82.6 & 44.6 & 42.4 & 62.0 & 68.6 & 44.8 & 58.5 \\
15 & 63.2 & 63.5 & 51.0 & 59.3 & 81.2 & 0.0 & 35.2 & 38.8 & 66.3 & 69.0 & 46.0 & 60.4 \\
16 & 58.5 & 69.9 & 55.9 & 61.4 & 99.8 & 100.0 & 3.6 & 67.8 & 59.2 & 86.9 & 48.8 & 65.0 \\
17 & 62.9 & 70.8 & 54.7 & 62.8 & 99.3 & 100.0 & 30.9 & 76.7 & 61.6 & 74.0 & 53.6 & 63.1 \\
18 & 67.3 & 75.9 & 61.3 & 68.2 & 99.5 & 70.0 & 47.0 & 72.2 & 62.6 & 94.7 & 62.5 & 73.3 \\
19 & 65.6 & 64.4 & 54.0 & 61.3 & 98.0 & 0.0 & 78.5 & 58.8 & 66.9 & 80.5 & 62.7 & 70.0 \\
20 & 67.8 & 66.0 & 47.5 & 60.4 & 97.6 & 0.0 & 0.8 & 32.8 & 65.5 & 85.8 & 54.7 & 68.6 \\
21 & 58.2 & 63.5 & 44.2 & 55.3 & 99.4 & 100.0 & 42.4 & 80.6 & 61.0 & 69.1 & 45.4 & 58.5 \\
22 & 62.8 & 57.6 & 58.2 & 59.5 & 99.8 & 98.0 & 79.5 & 92.4 & 64.5 & 70.8 & 59.7 & 65.0 \\
23 & 70.0 & 71.7 & 55.6 & 65.8 & 99.5 & 2.5 & 14.0 & 38.7 & 69.6 & 71.9 & 57.7 & 66.4 \\
24 & 72.3 & 63.0 & 53.5 & 62.9 & 0.0 & 0.3 & 46.0 & 15.4 & 68.8 & 73.7 & 67.2 & 69.9 \\
25 & 69.8 & 63.2 & 56.3 & 63.1 & 0.2 & 0.0 & 4.7 & 1.6 & 58.6 & 65.8 & 76.9 & 67.1 \\
26 & 65.1 & 70.2 & 56.4 & 63.9 & 9.4 & 94.2 & 56.5 & 53.4 & 64.4 & 74.5 & 67.7 & 68.8 \\
27 & 68.9 & 67.2 & 68.2 & 68.1 & 99.8 & 95.1 & 45.1 & 80.0 & 62.7 & 85.8 & 78.3 & 75.6 \\
28 & 57.8 & 58.9 & 61.5 & 59.4 & 17.9 & 99.9 & 94.9 & 70.9 & 63.8 & 76.2 & 86.1 & 75.4 \\
29 & 71.6 & 78.7 & 49.3 & 66.5 & 0.0 & 100.0 & 11.4 & 37.1 & 63.9 & 74.4 & 79.7 & 72.7 \\
30 & 59.5 & 59.6 & 57.2 & 58.8 & 0.0 & 100.0 & 39.2 & 46.4 & 67.8 & 68.6 & 72.9 & 69.8 \\
31 & 64.4 & 67.3 & 46.1 & 59.3 & 100.0 & 0.0 & 1.2 & 33.7 & 69.3 & 71.0 & 57.7 & 66.0 \\
\midrule
avg & 64.8 & 67.4 & 53.3 & — & 60.6 & 47.7 & 43.8 & — & 61.2 & 77.9 & 67.6 & — \\
\bottomrule
\end{tabular}
\caption{Module-wise average sparsity (\%) of LLaMA3.1-8B on commonsense reasoning. ``-'' denotes rank 0.}
\label{tab:layerwise}
\end{table*}
\subsection{Hyperparameters}
\label{apx:a4}
\begin{table*}[t]
\centering
\small
\setlength{\tabcolsep}{3pt}

\begin{tabular}{lcccccccccc}
\toprule
Task & Data & Rank & Scale & Dropout & Optimizer & LR & Scheduler & Batch & Epochs & Target \\
\midrule

CSR & ALL & 32 & 0.5 & 0.05 & AdamW & 1e-4 & Linear & 16 & 3 & Q,K,V,Up,Down \\
VIT & ALL & 128 & 0.5 & 0.05 & AdamW & 2e-4 & Cosine & 16 & 1 & Q,K,V,O,Up,Down,Gate \\
NLU & CoLA  & 4 & 0.5 & 0.1  & AdamW & 4e-4   & Linear & 32 & 40 & Q,K,V,O,I \\
NLU & MNLI  & 4 & 0.5 & 0.15 & AdamW & 2.5e-4 & Linear & 32 & 10 & Q,K,V,O,I \\
NLU & QNLI  & 4 & 0.5 & 0.1  & AdamW & 2e-4   & Linear & 32 & 10 & Q,K,V,O,I \\
NLU & RTE   & 4 & 0.5 & 0.2  & AdamW & 1e-4   & Linear & 32 & 80 & Q,K,V,O,I \\
NLU & MRPC  & 4 & 0.5 & 0    & AdamW & 1e-4   & Linear & 32 & 60 & Q,K,V,O,I \\
NLU & QQP   & 4 & 0.5 & 0.15 & AdamW & 4e-4   & Linear & 32 & 10 & Q,K,V,O,I \\
NLU & SST-2 & 4 & 0.5 & 0    & AdamW & 4e-4   & Linear & 32 & 48 & Q,K,V,O,I \\
NLU & STS-B & 4 & 0.5 & 0.2  & AdamW & 5e-4   & Linear & 32 & 45 & Q,K,V,O,I \\
NLG & ALL & 32 & 0.5 & 0.05 & AdamW & 2e-4 & Linear & 8 & 3 & Q,K,V,O,Up,Down,Gate \\
\bottomrule
\end{tabular}
\caption{Hyperparameter configuration for our experiments.}
\label{tab:hyper_all}
\end{table*}

Across all experiments, we tune the sparsity-related hyperparameters s and $\lambda$ separately from the other training configurations. For CSR and VIT, we fix the scaling factor at s = $4\times 10^{-5}$ and select $\lambda$ from the interval $[10^{-7}, 10^{-3}]$ based on development performance. For NLU tasks, we instead set $s = 1\times 10^{-4}$ and choose $\lambda$ from $[10^{-8}, 10^{-3}]$. For NLG task, we set $s = 4\times 10^{-5}$, and select $\lambda$ from $[10^{-6}, 10^{-3}]$. The remaining hyperparameters, including rank, scaling coefficient, dropout rate, optimizer, learning rate, scheduler, batch size, number of epochs, and target modules, are summarized in Table~\ref{tab:hyper_all}.

\subsection{Hyperparameter Sensitivity Analysis}

\begin{figure}[t]
    \centering
    \includegraphics[width=\linewidth]{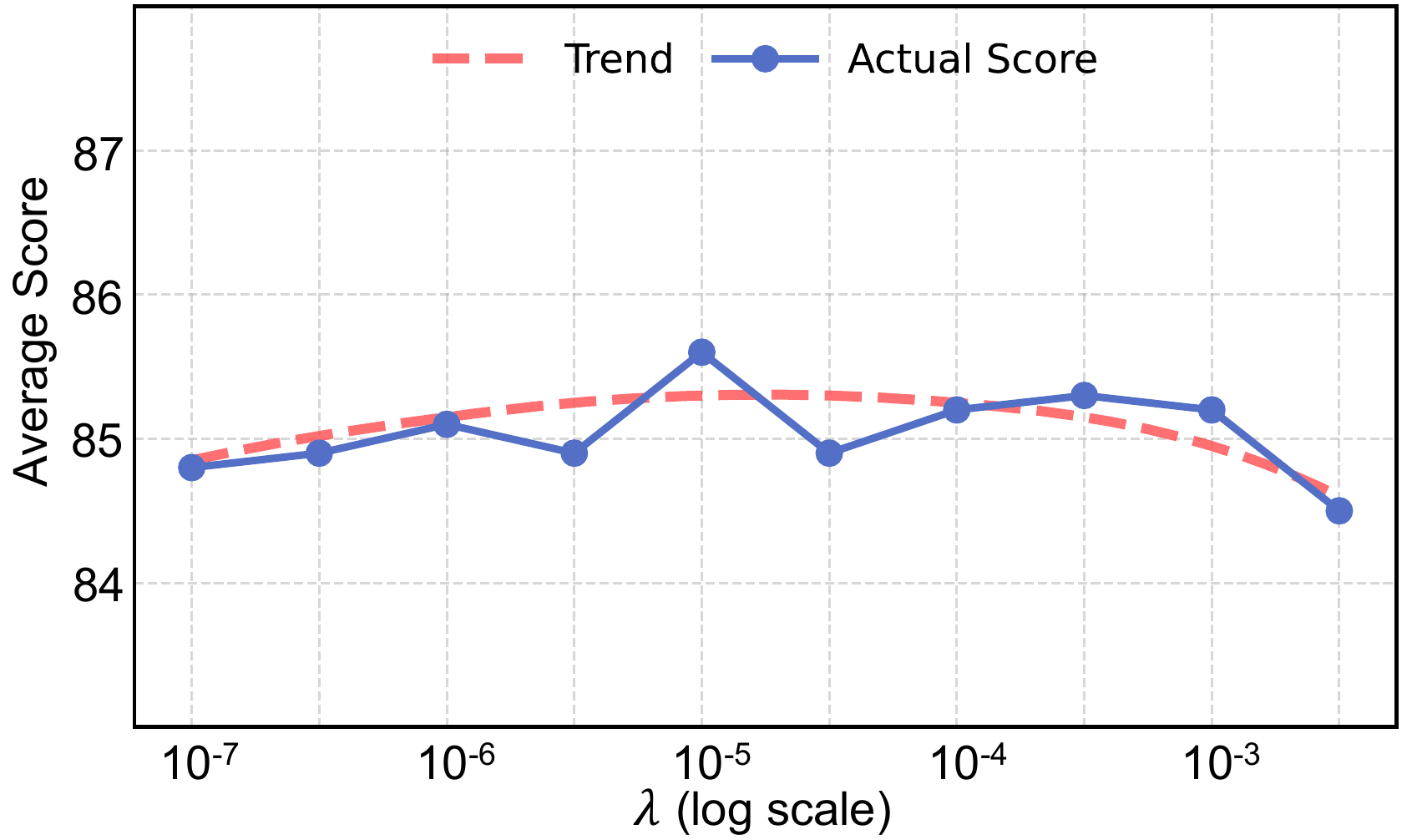}  
    \caption{Sensitivity analysis of the sparsity penalty term $\lambda$ for TS-DoRA on CSR benchmarks. The trend indicates that appropriate sparsity enhances performance by mitigating noise, whereas excessive sparsity impairs model capacity, thereby confirming the existence of an optimal level of redundancy.}
    \label{fig:lambda}
\end{figure}

\paragraph{Impact of the Sparsity Penalty $\lambda$.}
To investigate the sensitivity of TS-PEFT to the sparsity regularization term, we conduct a study on the hyperparameter $\lambda$ using the TS-DoRA model on CSR benchmarks. As defined in the objective function Eq. (4), $\lambda$ governs the trade-off between minimizing the task loss $l(\tau)$ and enforcing sparsity. Figure~\ref{fig:lambda} illustrates the trend of the Average Score on CSR tasks as $\lambda$ varies. We observe a distinct ``inverted U-shaped'' trend, which aligns well with our theoretical expectations:
\begin{itemize}
    \item \textbf{Low $\lambda$ Region (Left side of the curve)}: When $\lambda$ is extremely small, the penalty for activating tokens is negligible. Thus, the threshold $\tau$ remains at a low level, and the model behavior asymptotically converges to the standard, dense-update DoRA baseline. Although the performance is acceptable, it remains suboptimal due to the presence of redundant updates and optimization noise.
    \item \textbf{Optimal $\lambda$ Region (Peak)}: As $\lambda$ increases to the optimal range, the model effectively prunes redundant token updates. This peak performance strongly validates our insight: selectively skipping non-critical updates can mitigate noise and enhance generalization capability.
    \item \textbf{High $\lambda$ Region (Right side of the curve)}: When $\lambda$ is excessively large, the regularization term dominates the optimization objective. Even critical updates required for task adaptation are suppressed. This triggers underfitting and leads to a sharp decline in performance.
\end{itemize}
\paragraph{Role of the Gradient Scaling Factor $s$.} Regarding the hyperparameter $s$, we focus on its role within the threshold update rule defined in Eq. (20). Here, $s$ serves as a global scaling factor. Given that the derivative of the step function is theoretically zero almost everywhere, $s$ essentially determines the effective step size magnitude for learning $\tau$, functioning similarly to a Base Learning Rate. Empirically, we found that treating $s$ as a fixed constant suffices to achieve stable convergence. As shown in Appendix A.4, we set $s$ to $4 \times 10^{-5}$ for CSR, VIT and NLG tasks and $1 \times 10^{-4}$ for NLU tasks, relying on adaptive momentum estimation to manage dynamic adjustments during training.

\end{document}